\documentclass[conference]{IEEEtran}
\IEEEoverridecommandlockouts
\usepackage{cite}
\usepackage{amsmath,amssymb,amsfonts}
\usepackage{algorithmic}
\usepackage{graphicx}
\usepackage{textcomp}
\usepackage{xcolor}
\def\BibTeX{{\rm B\kern-.05em{\sc i\kern-.025em b}\kern-.08em
    T\kern-.1667em\lower.7ex\hbox{E}\kern-.125emX}}
\begin{document}

\title{Alzheimer's Dementia Detection using Acoustic \& Linguistic features and Pre-Trained BERT\\

\thanks{This work was funded by the Department of Science and
technology, Government of India, vide Reference No:
SR/CSI/50/2014 (G) through the Cognitive Science Research
Initiative (CSRI). We acknowledge financial supports from the EU: ERC-2017-COG project IONOS (\# GA 773228). }
}

\author{\IEEEauthorblockN{Akshay Valsaraj}
\IEEEauthorblockA{\textit{Cognitive Neuroscience Lab} \\
\textit{BITS Pilani, K.K. Birla Goa Campus}\\
Goa, India \\
f20180608@goa.bits-pilani.ac.in}
\and
\IEEEauthorblockN{Ithihas Madala}
\IEEEauthorblockA{\textit{Cognitive Neuroscience Lab} \\
\textit{BITS Pilani, K.K. Birla Goa Campus}\\
Goa, India \\
f20180607@goa.bits-pilani.ac.in}
\and
\IEEEauthorblockN{Nikhil Garg}
\IEEEauthorblockA{\textit{IEMN} \\
\textit{ Université de Lille}\\
59650 Villeneuve d’Ascq, France \\
Nikhil.Garg@Usherbrooke.ca}
\and
\IEEEauthorblockN{Veeky Baths}
\IEEEauthorblockA{\textit{Cognitive Neuroscience Lab} \\
\textit{BITS Pilani, K.K. Birla Goa Campus}\\
Goa, India \\
veeky@goa.bits-pilani.ac.in}
}

\maketitle

\begin{abstract}

Alzheimer's disease is a fatal progressive brain disorder that worsens with time. It is high time we have inexpensive and quick clinical diagnostic techniques for early detection and care. In previous studies, various Machine Learning techniques and Pre-trained Deep Learning models have been used in conjunction with the extraction of various acoustic and linguistic features. Our study focuses on three models for the classification task in the  ADReSS  (The  Alzheimer’s  Dementia  Recognition through  Spontaneous  Speech)  2021  Challenge. We use the well-balanced dataset provided by the ADReSS Challenge for training and validating our models. Model 1 uses various acoustic features from the eGeMAPs feature-set, Model 2 uses various linguistic features that we generated from auto-generated transcripts and Model 3 uses the auto-generated transcripts directly to extract features using a Pre-trained BERT and TF-IDF. These models are described in detail in the models section.\\

\end{abstract}

\begin{IEEEkeywords}
Speech Recognition, Human-Computer-Interaction, Computational Para-linguistics
\end{IEEEkeywords}

\section{Introduction}

Alzheimer’s Dementia (AD) is a progressive Neuro-degenerative disorder that affects major brain functions. The usual manifestations of AD include visible changes in memory and language and cognitive function. It also leads to personality changes along with agitation and losing the train of thought while speaking. Unlike other illnesses, the onset of visible symptoms is not observed in the initial stages. Early detection of this disease can prevent further degeneration, improve life quality and life expectancy.

Hence, AD is a major health concern in the struggle for increased life expectancy in the general population. Initially, it was considered to be a rare disease but soon it became closely associated with aging. There is a lot of stigma attached with this disease but with increased research and new treatments for patients with AD, these misconceptions are slowly fading away. Hence, it is the need of the hour to come up with inexpensive, quick and accurate diagnostic tools and early detection methods to provide utmost care to people suffering from early stages of the disease.

Current methods of diagnosis are human intensive, lab intensive and hence expensive and time consuming. While traditional Brain Imaging techniques are effective in detecting the onset of disease, these are expensive and tedious to repeatedly perform on a patient or a healthy individual. Microphones are present in most laptops and smartphones, and most people possess at least one of the two. Thus, speech data could be easily collected from any healthy individual without requiring any medical intervention. This spontaneous speech data could be used for early detection of AD when setup in the form of an interview taken by a virtual assistant and recording the responses.

The comparison of previously proposed models (ADReSS 2021 Challenge) for the classification task is summarized in Table I. This includes the features extracted, ML/DL models used and Accuracy achieved on the test dataset.

In our study, we discuss three different models based on the input features. Model 1: The Acoustic Model, uses various acoustic features from the eGeMAPs feature-set, Model 2: The Lingustic Model, uses various linguistic features that we generated from auto-generated transcripts and Model 3: Pre-trained BERT model + TF-IDF, uses the auto-generated transcripts directly to extract features using a Pre-trained BERT and TF-IDF. 

We also propose the implementation of models based on Vision Transformers (ViT) and sPCEN normalized spectrograms as future work. 

\begin{table}[htbp]
\caption{Speech classification models of Interspeech 2020}
\centering
\resizebox{\columnwidth}{!}{%
\begin{tabular}{|l|l|l|l|}
\hline
\multicolumn{1}{|c|}{Input features}                                                                                   & \multicolumn{1}{c|}{Models}                                                                     & \multicolumn{1}{c|}{Accuracy} & \multicolumn{1}{c|}{Rf} \\ \hline
\begin{tabular}[c]{@{}l@{}}Extracted \\ voice features\\ (acoustic) and \\ transcripts\end{tabular}                   & \begin{tabular}[c]{@{}l@{}}LDA,DT,\\ INN,SVM,\\ RF\end{tabular}                                      & 75\%                          &   \cite{Luz}                      \\ \hline
\begin{tabular}[c]{@{}l@{}}Audio ,\\ TF-IDF and\\ readability \\ features, \\ embeddings.\end{tabular}                & \begin{tabular}[c]{@{}l@{}}Xgboost, \\ SVM, \\ RF and LR\end{tabular}                                & 77.8\%                        & \cite{Martinc}   \\ \hline
\begin{tabular}[c]{@{}l@{}}Fusion of \\ Acoustic and \\ silence features \\ with transcripts\end{tabular}              & \begin{tabular}[c]{@{}l@{}}x-vectors ,\\  BERT\end{tabular}                                          & 75\%                          & \cite{Pappagari}  \\ \hline
\begin{tabular}[c]{@{}l@{}}Acoustic \\ and lexical \\ information, \\ including \\ disfluency \\ tagging\end{tabular} & \begin{tabular}[c]{@{}l@{}}LSTM with \\ Gating\end{tabular}                                          & 79\%                          & \cite{Rohanian} \\ \hline
\begin{tabular}[c]{@{}l@{}}Phonemes \\ and Audio\end{tabular}                                                         & \begin{tabular}[c]{@{}l@{}}FastText \\ supervised \\ classifier\end{tabular}                         & 79.1\%                        & \cite{Edwards} \\ \hline
\begin{tabular}[c]{@{}l@{}}Textual \\ Transcriptions\end{tabular}                                                     & \begin{tabular}[c]{@{}l@{}}PAR+INV / \\ DistilBERT\end{tabular}                                      & 81\%                          & \cite{Searle} \\ \hline
\begin{tabular}[c]{@{}l@{}}Speaker \\ Representation \\ Vectors and \\ Textual Feature \\ Embeddings\end{tabular}     & \begin{tabular}[c]{@{}l@{}}pre-trained \\ x-vectors SRE\\ and\\  LSTM-RNNs\end{tabular}              & 81.25\%                       & \cite{Pompili} \\ \hline
\begin{tabular}[c]{@{}l@{}}Chat Transcipts ,\\ Acoustic and  \\ Semantic features\end{tabular}                       & \begin{tabular}[c]{@{}l@{}}Pre-trained \\ BERT and \\ SVM\end{tabular}                               & 83.3\%                        &  \cite{Balagopalan}               \\ \hline
\begin{tabular}[c]{@{}l@{}}Audio \\ Features and \\ Transcripts \end{tabular}                                           & \begin{tabular}[c]{@{}l@{}}Bag-of-\\ Audio-Words, \\ Siamese and \\ Attention\\ Network\end{tabular} & 85.2  \%                      & \cite{Cummins} \\ \hline
Chat Transcripts                                                                                                      & \begin{tabular}[c]{@{}l@{}}Pre-trained \\ BERT and \\ ERNIE models\end{tabular}                      & 89.6\%                        & \cite{Yuan} \\ \hline
\begin{tabular}[c]{@{}l@{}}Disfluency,\\ acoustic and \\ Interventions \\ features\end{tabular}                       & \begin{tabular}[c]{@{}l@{}}MLP ,RNN \\ and ensemble\\  models\end{tabular}                           & 83\%                          & \cite{Sarawgi} \\ \hline
\begin{tabular}[c]{@{}l@{}}Pre-trained \\ language \\ model and \\ linguistic \\ features.\end{tabular}               & \begin{tabular}[c]{@{}l@{}}(bi-LSTM) \\ and \\ CNN\end{tabular}                                      & 81.25\%                       & \cite{Koo} \\ \hline

\begin{tabular}[c]{@{}l@{}}Acoustic \\ features and\\  Speech\\ Transcripts\end{tabular}                             & \begin{tabular}[c]{@{}l@{}}SVM + \\ Bert features\end{tabular}                                       & 85.42\%                       & \cite{Syed} \\ \hline

\end{tabular}}
\end{table}


\section{Dataset}
We use the ADReSS Challenge Dataset which consists of 242 audio files. These were split into training and test sets with 70\% for training and 30\% for testing and validation. A set of recordings of picture descriptions produced by cognitively normal subjects and patients with an AD diagnosis, who were asked to describe the Cookie Theft picture from the Boston Diagnostic Aphasia Exam.
The data set for the task was created from a study involving AD patients. The task involved classifying patients into 'ad' (suffering from Alzheimer's) and 'cn' (control subjects).
From the data set provided we removed the voice of the interviewer and kept only the voice of the patient using the utterance segmentation files (diarization) provided with the dataset.

\section{Transcripts}
For our linguistic model, we require auto-generated transcripts of each audio file. We used the PyTorch based pre-trained enterprise-grade Silero STT (Speech-To-Text) English (en\_v2) model for its ease of use \& integration with our code and it is also comparable to the quality of Google's STT according to the benchmarks.

\section{Feature extraction}
We calculated various linguistic features such as Brunet’s Index, Honore’s Statistic, Standardised Entropy, Root Type-Token Ratio and Hyper Geometric Distribution Diversity for our linguistic model. We extracted the eGeMAPS acoustic feature-set for our Acoustic Model. These features have shown to be very effective in previous studies related to Alzheimer's Dementia.
\subsection{Linguistic features}
Using the auto-generated transcripts, we decided to extract the following linguistic features based on extensive literature review.

\subsubsection{Brunet’s Index}

Its a measure of lexical diversity that has been used in analysis of text and its often considered to be not dependent on text length. Its calculated by using the formula:
\begin{align}
		W=N^{V^{-a}}
\end{align}
where $N$ is the text length, $V$ is the number of different words and $a$ is a scaling constant which is usually set at $0.172$
The richer the vocabulary the lower the values of $W$ and the values usually range between $10$ and $20$. Patients suffering from dementia tend to repeat their words which leads to a lower lexical diversity, hence an important feature to consider.

\subsubsection{Honore’s Statistic}
This feature has been used in the stylometric analyses of texts. It is based on the fact that texts which have a higher proportion of words tend to have a richer vocabulary. Its calculated by using the formula:
\begin{align}
R=\dfrac{100\log \left( N\right) }{\left( 1-\dfrac{v_{1}}{V}\right) }
\end{align}
where $v_{1}$ is the number of words used only one, $V$ is the number of different words and $N$ is the text length. Higher values of R correlate to a richer vocabulary and lower values correlate to poorer vocabulary. Patients suffering from dementia tend to have depleting vocabulary, hence an important feature to consider.

\subsubsection{Standardised Entropy} 
This measure helps us to check the degradation in the variety of words and word combinations that a patient uses. It is calculated by dividing the word entropy by the log of the total word count. Formula:
\begin{align}
\dfrac{-\sum P\left( xi\right) \times \log _{2}P\left(xi\right) }{\log \left( N\right)}
\end{align}
where $P(xi)$ is the probability of the word in the text and $N$ is the total number of words.

\subsubsection{Root Type-Token Ratio}
It has been noticed that in patients who suffer from Alzheimer's, there is a lot of repetition of words in their conversation as they tend to forget what they said and tend to repeat it again. Hence, to measure this we use Type-Token Ratio and to reduce its variance on text length Root Type-Token ratio is used. Its calculated by using the formula:

\begin{align}
RTTR=\dfrac{t}{\sqrt{n}}
\end{align}
where $t$ is the number of unique words and $n$ is the total number of words.

\subsubsection{Mean Segment Type-Token Ratio} 
In this process the text to be analysed is divided into equal segments in terms of the number of words, in this case we used 16 words per segment.After this the TTR is calculated for each segment and the MSTTR is obtained after we take an arithmetic mean of all the TTR.

\subsubsection{Measure of Textual Lexical Diversity (MTLD)} 
This measure is similar to MSTTR but in this case the text is varaible and the number of words in each segment varies and the segment ends when the TTR reaches a threshold of 0.72.After that the following formula is obtained to obtain MLTD.
\begin{align}
MLTD=\dfrac{L}{n}
\end{align}
where $L$ is text length in number of words and $n$ is
the number of segments.

\subsubsection{Hyper Geometric Distribution Diversity} 
This index calculates the probability of finding any token for for each lexical type in a given text, in a random sample of 42 words taken from the same text. The sum of all the probabilities for all lexical types gives the final HDD score.

\subsubsection{POS-based features} 
We used this marker as tagging the text with POS tags can indicate frequent adjectives, verbs and adverbs which indicate more descriptive and meaningful sentences, while frequent adverbs can indicate the ability to relate different utterances to each other. Hence, it can be a useful identifier of patients with Alzheimer’s disease. The various POS tags we used are - \\
\begin{itemize}
\item Verb frequency: It is computed by dividing the total number of words with that verb tag by the total number of words spoken by the patient in the recording.
\item Noun frequency: It is computed by dividing the total number of words with that verb tag by the total number of words spoken by the patient in the recording.
\item Pronoun frequency: It is computed by dividing the total number of words with that Pronoun tag by the total number of words spoken by the patient in the recording.
\item Adverb frequency: It is computed by dividing the total number of words with that Adverb tag by the total number of words spoken by the patient in the recording.
\item Adjective frequency: It is computed by dividing the total number of words with that Adjective tag by the total number of words spoken by the patient in the recording.
\end{itemize}

\subsection{Acoustic Features}
For the Acoustic Model, we extracted the eGeMAPS feature set on the entire audio file using the $opensmile$ python package. The eGeMAPS feature set is a collection of 88 acoustic features that provide a baseline for evaluation in researches to avoid differences of feature set and implementations. We used a Recursive Feature Elimination (RFE) method to reduce the features which uses an external estimator that assigns weights to features and the RFE is used to select features by recursively considering smaller and smaller sets of feature.
By using this technique We found that only 51 features contribute more towards the prediction, hence we decided to proceed with them for training our model.

\section{Classification}
We used the following ML algorithms as final layers of our models -

\subsection{ Support Vector Machine(SVM)}
Support vector machines are one of the most prominent and widely used machine learning algorithms used for password authentication and accuracy prediction \cite{manian, yma, abdel}. SVMs are based on the idea that the classes can be separated linearly using a hyper-plane to maximize the spaces between the most classes. In non-linear classification, a kernel function is used to map the non-linear classes to a linear problem by raising their dimension. The most widely used kernel functions are linear, polynomial, radial-basis function (RBF), and sigmoidal. In this study, the Polynomial kernel (degree=4) was found to have better performance than the other kernels.

\subsection{Logistic Regression}
It is  a supervised Machine Learning algorithm that often uses a Sigmoid function to model a binary dependent variable. Instead of fitting a straight line or hyper-plane in the case of other model,s the logistic regression model uses a logistic function to predict the output of a linear equation between 0 and 1.

\subsection{Random forests}
Random Forest is a supervised Machine Learning algorithm \cite{rn} used for classification and regression tasks. It creates multiple decision trees by taking various combinations of classes and samples from the training set. Decision trees are similar to flow charts in terms of structure. A node represents a test on an attribute, each branch of the node holds an outcome of the test, and each leaf node is a class label. The root node is at the topmost position. The impurity in the decision trees constructed is reduced by calculating the Gini index and using it in performing the appropriate splitting measures. The Gini index is calculated using the formula:
\begin{align}
Gini =  1 - \displaystyle\sum_{i=1}^{n}(P_i)^2
\end{align}
where \(P_i\) is the probability of a node being classified into a particular class.

\section{Models}
We have three models - Model 1: The Acoustic Model, uses various acoustic features from the eGeMAPs feature-set, Model 2: The Lingustic Model, uses various linguistic features that we generated from auto-generated transcripts and Model 3: Pre-trained BERT model + TF-IDF, uses the auto-generated transcripts directly to extract features using a Pre-trained BERT and TF-IDF.

\subsection{Model 1: The Acoustic model}
In this model we used the eGeMAPS feature-set and obtained the final 51 features after feature reduction using Recursive Feature Elimination. We then classified them using various machine learning models such as Support Vector Machines (SVM), Linear Regression (LR) and Random Forest (RF).

\subsection{Model 2: The Linguistic model}
In this model we used the 13 features obtained from the various linguistic features mentioned in the previous section and classified them using the same machine learning models.

\subsection{Model 3: Pre-trained BERT model + TF-IDF}
Attention-based models \cite{Vaswani} belong to a class of AI models, commonly referred to as sequence-to-sequence models. The goal of these models, as the name suggests, is to generate an output sequence from an input sequence that is generally of different lengths. 
The main technical innovation of bidirectional transformers (BERT) \cite{Devlin} is to apply the bidirectional training of a transformer, a popular attention model, to language modelling. Using this approach, we fine-tune the pre-trained BERT model by using the auto-generated transcripts of the audio files. We used the BERT model to obtain the pre-trained word embeddings for each sentence and obtained a total of 768 features and then we concatenated the BERT features with the TF-IDF features obtained from the text. TF-IDF is an information retrieval technique that weighs a term’s frequency (TF) in the text and its inverse document frequency (IDF). A TF and an IDF value is obtained from each word and then they are multiplied. The formula to calculate TF-IDF for each word:
\begin{align}
tfidf_{t,d} =  tf_{t,d}\cdot \log \frac{N}{df_t}
\end{align}

where $ tf_{t,d}$ is the number of times the term $t$ occurs in the text $d$ and ${df_t}$ is the total number of documents containing the term t and $N$ is the total number of documents.
After obtaining all the features we classified them using the same machine learning models - SVM, LR, RF.

\section{Results}

The 5-Fold Cross Validation results of Models 1, 2 and 3 is summarized in Table 2 below. Both LR and SVM performed similarly considering the CV Accuracy achieved. But, considering the Precision, Recall, Specificity and F1 Score, LR came out on top with more consistency and hence we went with LR for all our models for the classification.

\begin{table}[htbp]
\centering
\caption{5-Fold Cross Validation results of Models 1, 2 and 3}
\resizebox{\columnwidth}{!}{%
\begin{tabular}{|l|l|l|l|l|l|}
\hline
\textbf{Class} & \textbf{CV Accuracy} & \textbf{Precision} & \textbf{Recall} & \textbf{Specificity} & \textbf{F1 Score} \\ \hline
\multicolumn{6}{|l|}{\textbf{Model 1}}                                                                                    \\ \hline
LR             & \textbf{0.640}               & \textbf{0.645}              & 0.741           & \textbf{0.522}                & 0.690             \\ \hline
RF             & 0.580                 & 0.607              & 0.630            & \textbf{0.522}                & 0.618             \\ \hline
SVM            & \textbf{0.640}                 & 0.629              & \textbf{0.815}          & 0.435                & \textbf{0.710}              \\ \hline
\multicolumn{6}{|l|}{\textbf{Model 2}}                                                                                    \\ \hline
LR             & \textbf{0.660}                  & \textbf{0.867}              & \textbf{0.464}           & \textbf{0.909}                & \textbf{0.605}             \\ \hline
RF             & 0.620                 & 0.800                & 0.429           & 0.864                & 0.558             \\ \hline
SVM            & 0.620                 & 0.846              & 0.393           & \textbf{0.909}                & 0.537             \\ \hline
\multicolumn{6}{|l|}{\textbf{Model 3}}                                                                                    \\ \hline
LR             & \textbf{0.700}                  & \textbf{0.731}              & \textbf{0.704}           & 0.696                & \textbf{0.717}             \\ \hline
RF             & 0.618                & 0.562              & 0.600             & 0.632                & 0.581             \\ \hline
SVM            & \textbf{0.700}                  & 0.667              & 0.636           & \textbf{0.75}                 & 0.651            \\ \hline
\end{tabular}%
}
\end{table}

In Table 3, we have summarized the Test Set results. Clearly the Model 3 with the Pre-trained BERT + TF-IDF features is outperforming the Acoustic and Linguistic models. This is inline with the previous studies where Deep Learning based linguistic models outperform basic ML based models and acoustic models.

\begin{table}[htbp]
\caption{Test Set Results}
\resizebox{\columnwidth}{!}{%
\begin{tabular}{|l|l|l|l|l|l|}
\hline
Model  & Class                         & Accuracy    & Recall                                                  & Precision                                               & F1                                                      \\ \hline
Model 1 & \begin{tabular}[c]{@{}c@{}}non-AD\\ AD\end{tabular} & 0.595  & \begin{tabular}[c]{@{}l@{}}0.5000\\ 0.6857\end{tabular} & \begin{tabular}[c]{@{}l@{}}0.6207\\ 0.5714\end{tabular} & \begin{tabular}[c]{@{}l@{}}0.5538\\ 0.6234\end{tabular} \\ \hline
Model 2 & \begin{tabular}[c]{@{}c@{}}non-AD\\ AD\end{tabular} & 0.5775 & \begin{tabular}[c]{@{}l@{}}0.7222\\ 0.4286\end{tabular} & \begin{tabular}[c]{@{}l@{}}0.5652\\ 0.6000\end{tabular} & \begin{tabular}[c]{@{}l@{}}0.6341\\ 0.5000\end{tabular} \\ \hline
Model 3 & \begin{tabular}[c]{@{}c@{}}non-AD\\ AD\end{tabular} & 0.7606 & \begin{tabular}[c]{@{}l@{}}0.7222\\ 0.8000\end{tabular} & \begin{tabular}[c]{@{}l@{}}0.7879\\ 0.7368\end{tabular} & \begin{tabular}[c]{@{}l@{}}0.7356\\ 0.7671\end{tabular} \\ \hline
\end{tabular}%
}
\end{table}

\section{Future Work}

\subsection{Deep Acoustic Model: Vision Transformer and sPCEN}
Vision Transformer (ViT) \cite{vit} is now the state-of-the-art architecture for image classification tasks. The first layer of the Vision Transformer proposed in the paper involves segmentation of images into small patches (16x16). These patches are then linearly embedded, followed by the addition of position embeddings and then fed into a standard Transformer Encoder. The final classification token was also tweaked according the number of classes required for the task. We worked on implementing ViT for AD classification by constructing mel-spectrograms and fine-tuning pre-trained ViT, but we were not able to achieve respectable results. 

We propose to tweak the first layer such that we take a segment of dimension 128x16 which keeps the full height of the spectrogram (assuming the height of the spectrogram is 128 pixels and width is a multiple of 16) and train the model from scratch on the ADReSS dataset. This way of segmentation may improve the learnability of the model specifically for spectrograms, due to the way spectrograms are constructed. Previous studies have shown that breaking the audio files into 3-second clips and then constructing spectrograms separately for each clip improves accuracy. 

We also propose to normalize spectrograms with sPCEN \cite{Lostanlen}. A typical spectrogram uses linear frequency scaling so that each frequency range is the same number of Hertz apart, shown to outperform the point-wise logarithm of mel-frequency spectrogram as an acoustic front-end. However, in the case of PCEN (Per-Channel Energy Normalization), it improves the robustness of the channel distortion. It has the ability to Gaussianize and whiten the background of acoustic recordings which is due to the  operations of temporal integration, adaptive gain control, and dynamic range compression. Hence it is shown to outperform the point wise logarithm of Mel-frequency spectrogram.

\subsection{Ensemble model}
We would like to explore further by combining the Deep Acoustic model with our Model 3 (Pre-trained BERT) to find out if it improves the prediction accuracy.


\vspace{12pt}

\end{document}